\documentclass{article}
\usepackage{ijcai24}

\usepackage{times}
\usepackage{soul}
\usepackage{url}
\usepackage[hidelinks]{hyperref}
\usepackage[utf8]{inputenc}
\usepackage[small]{caption}
\usepackage{graphicx}
\usepackage{amsmath}
\usepackage{amsthm}
\usepackage{booktabs}
\usepackage{algorithm}
\usepackage{algorithmic}
\usepackage[switch]{lineno}
\usepackage{subcaption}
\usepackage{colortbl}
\usepackage{tabularx}

\title{MoVL:Exploring Fusion Strategies for the Domain-Adaptive Application of Pretrained Models in Medical Imaging Tasks}

\author{
Haijiang Tian$^1$
\and
Jingkun Yue$^1$\and
Xiaohong Liu$^{2}$\and
Guoxing Yang$^1$\and
Zeyu Jiang$^1$\and
Guangyu Wang$^{1,*}$\\
\affiliations
$^1$State Key Laboratory of Networking and Switching Technology, Beijing University of Posts and Telecommunications, Beijing, China\\
$^2$UCL Cancer Institute, University College London, London, United Kingdom\\
\emails
$^*$Corresponding Author: guangyu.wang@bupt.edu.cn
}

\begin{document}

\maketitle

\begin{abstract}
    Medical images are often more difficult to acquire than natural images due to the specialism of the equipment and technology, which leads to less medical image datasets. So it is hard to train a strong pretrained medical vision model. How to make the best of natural pretrained vision model and adapt in medical domain still pends. For image classification, a popular method is linear probe (LP). However, LP only considers the output after feature extraction. Yet, there exists a gap between input medical images and natural pretrained vision model. We introduce visual prompting (VP) to fill in the gap, and analyze the strategies of coupling between LP and VP. We design a joint learning loss function containing categorisation loss and discrepancy loss, which describe the variance of prompted and plain images, naming this joint training strategy MoVL (\textbf{M}ixture \textbf{o}f \textbf{V}isual Prompting and \textbf{L}inear Probe). We experiment on 4 medical image classification datasets, with two mainstream architectures, ResNet and CLIP. Results shows that without changing the parameters and architecture of backbone model and with less parameters, there is potential for MoVL to achieve full finetune (FF) accuracy (on four medical datasets, average 90.91\% for MoVL and 91.13\% for FF). On out of distribution medical dataset, our method(90.33\%) can outperform FF (85.15\%) with absolute 5.18 \% lead.
\end{abstract}

\section{Introduction}


Recently, the evolution of foundation models in deep learning represent the paradigm of pre-training followed by finetuning. These models are extensively trained on tremendous data, establishing a broad base of knowledge, and then adapted to specific downstream tasks through finetuning. However, it is challenging for specific domains such as medical images, where the collection of substantial medical data is limited by privacy concerns or other constraints. Unlike the ease of capturing natural data, acquiring medical images such as X-ray is not as straight forward, leading to  a scarcity of medical data compared to natural data. 
Despite effortes in \cite{wang2022medclip,zhou2022self} developing visual pretrained models in specific medical domains, their performance often cannot match that of natural models. A usual and reasonable way is to make best use of natural pretrained model, which contains immense knowledge for feature extraction, to adapt in medical downstream tasks. While medical images and natural images differ significantly, they share certain features such as edge, corner and point. However, how to efficiently adapting the natural pretrained models for medical downstream tasks remains an unresolved challenges.

In addressing the challenge of adapting pre-trained models for downstream tasks, a commonly considered approach is full finetuning the pretrained model. However, this method needs to update numerous parameters, leading to considerable computation cost. In addition, there are some parameter efficient finetuning methods, such as Adapter \cite{chen2022vision} and LoRA \cite{hu2021lora}. However, most of them require model modifications. Another popular adaptation method is linear probe, which adds a fully connected layer after the image encoder, does not change the original architecture with a minimal set of trainable parameters. Yet, as illustrated in Figure \ref{figure2}, a significant gap exists between the input distribution of the medical images and natural pretrained model. This discrepancy raises a crucial question of input for LP.

In natural language processing (NLP), prompt is an emerging method to adapt downstream tasks \cite{petroni2019language,brown2020language,wei2022chain,hambardzumyan2021warp}. For example, give the prompt "Suppose you play a role of a programmer" to guide the large language model (LLM) to solve the problems in coding. Prompt is considered to change the input distribution so that being closer to the distribution of downstream tasks \cite{liu2023pre}. Inspired by this idea, visual prompt (VP) is introduced to computer vision tasks. Prompting a point or a box to stress on the object for segmentation may be intuitive \cite{kirillov2023segment}. By adding prior knowledge, we can lead the model to focus on specific location. Another method tries to add perturbation to the input images, which changes the distribution of original images \cite{bahng2022exploring}. This process may fill the gap between medical images and natural pretrained model.

While this method's potential lies in its orthogonal to other finetuning methods that shift the input distribution, it still faces the issue of label matching(LM). In scenarios where classifiers like ResNet are not specifically trained, we may establish rules to matching the labels. For instance, one may select the first ten categories from the 1000 Imagenet \cite{deng2009imagenet} categories for CIFAR10 \cite{krizhevsky2009learning} classification task. Although there are existing work on LM \cite{chen2023understanding}, it still challenging to fill the gap of output without training LP. For CLIP \cite{radford2021learning}, it seems getting a relatively better results because of text prompt, but when it switched to medical downstream tasks, complicated terms can also be difficult to extract text information. On account of output label matching limitation, how to fill the output gap for VP?

\begin{figure}[t]
    \centering
    \includegraphics[width=0.5\textwidth]{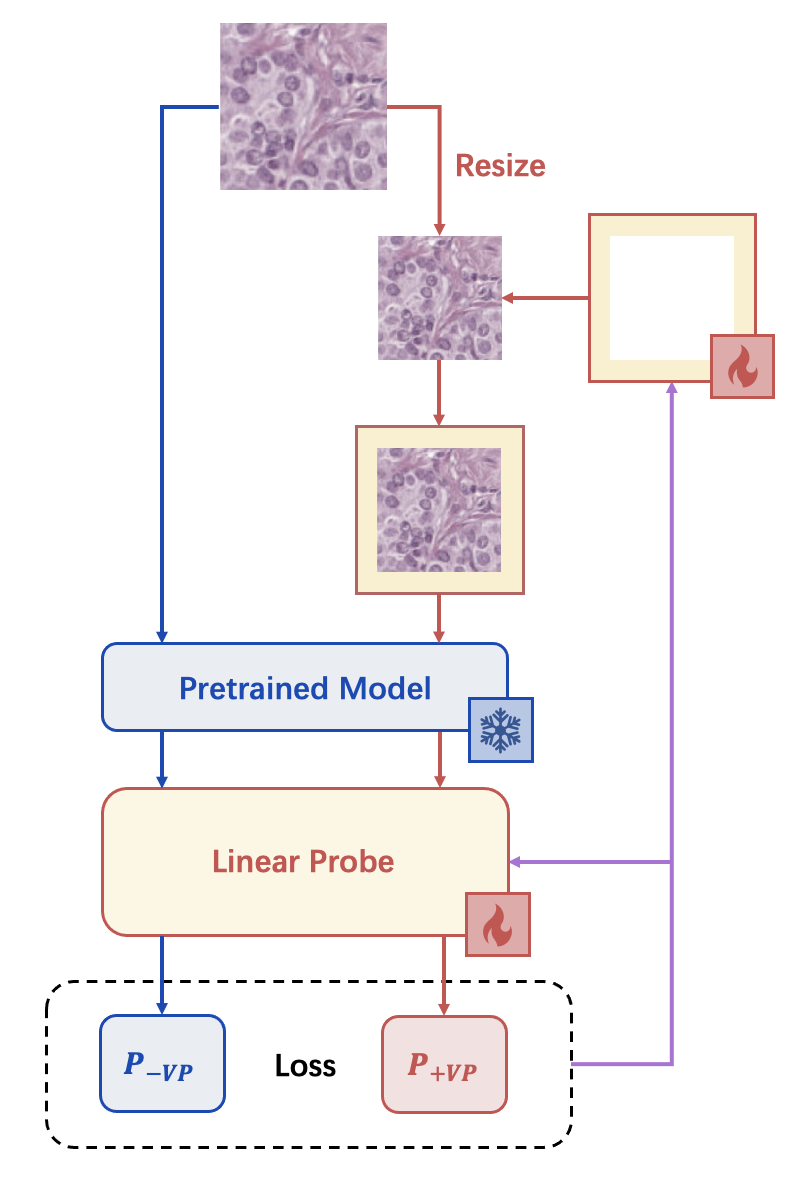}
    \caption{The overview of MoVL. VP and LP are trainable, and pretrained model is frozen. $P_{-VP}$ and $P_{+VP}$ are both computed in loss, and use back forward propogation to update LP and VP. $P_{-VP}$ is detached from computational graph. Red and blue lines are forward process, purple lines show the backward propogation direction.}
    \label{figure1}
\end{figure}

To tackle the previous challenges, we propose a method, named Mixture of Visual Prompting and Linear Probe (MoVL), which combines the adaptability of VP with the strength of LP for transfer learning, effectively addressing both input and output gaps without modifying the architecture or parameters of the pretrained model. We take different training strategies into consideration and choose an efficient mixed strategy, which trains VP and LP together from the initial stage. As shown in Figure \ref{figure1}, our method also considers the inherent information in the original image without VP . The original output is used as a reference in the loss function, enhancing the prompted output. Our contributions are listed below:


1. We propose an innovative mixed training strategy that effectively combines the adaptability of Visual Prompting (VP) with the strength of Linear Probe (LP) in transfer learning. This strategy commences from the initial training phase, simultaneously training both VP and LP components, ensuring a harmonized integration of their benefits.


2. Based on the mixed strategy, we design a novel joint learning loss comprising categorization loss and discrepancy loss, to utilize information from the original image. It updates model parameters by contrasting original images with their prompted counterparts, thus improving classification accuracy.

3. We perform extended experiments on different architectures and sizes of backbone models. According to the results, our strategy and loss can improve the accuracy on different medical downstream classification datasets.

\section{Related Work}

\subsection{Prompt in NLP}
In Natural Language Processing (NLP), the technique of prompting has been extensively utilized to stimulate and guide Pretrained Language Models (PLMs) in downstream tasks, which can be categorized into two types: hard prompts and soft prompts.

Hard Prompts, also known as templates, typically refer to predefined and fixed input patterns that are directly inserted into the model's inputs during model training or inference. \cite{petroni2019language} effectively demonstrated the potential of hard prompts in extracting implicit knowledge from pre-trained models. \cite{brown2020language} revealed the important role of hard prompts in quickly adapting models to new tasks. \cite{wei2022chain} utilised "chain of thought" to enhance the model's capability in logical reasoning and problem solving.

Soft Prompts, on the other hand, are those prompts that are dynamically, which do not change the parameters of the model, but rather influence the model's behaviour by adjusting the distribution of inputs. \cite{hambardzumyan2021warp} discussed that soft prompts can improve the flexiblity and effectiveness of pre-trained language models. \cite{chen2024eliciting} demonstrated the effectiveness of continuous soft prompts in enhancing the model's ability to perform knowledge-intensive tasks, whilst also highlighting the potential of automatically generated prompts in reducing manual intervention and increasing efficiency.

From the above research, it is evident that prompting techniques show immense potential in enhancing language model performance, knowledge extraction, and understanding model behavior.

\subsection{Prompt in CV}

As computer vision (CV) advancing, prompting techniques have expanded into visual tasks. While not yet widely applied, recent studies \cite{zhou2022learning,gu2023systematic} have demonstrated the unique potential of visual prompting techniques in enhancing model performance, interpreting model behavior, and facilitating cross-modal understanding.

A study \cite{jia2022visual} demonstrated the effectiveness of visual prompt tuning (VPT) on a variety of image classification tasks. The approach achieves prompting of input visual features by adding a small number of learnable parameters before the transformer block. This approach adds trainable prompts only at the input of the model, avoiding the need to completely retrain the model and effectively reducing the training cost while maintaining the performance of the model. 

Visual prompting \cite{bahng2022exploring}, by adding a ring of learnable perturbation pixels around the input images, guides the model to better perform in specific visual downstream tasks. Activation Prompt \cite{anonymous2023visual} involves adding perturbations to the feature maps generated during the feature extraction process. This method enhances the training efficiency of visual prompts and offers a new perspective for understanding and improving the internal workings of visual models. Another research \cite{chen2023understanding}, from the perspective of label mapping, explored and optimized the correspondence between visual prompts and labels to improve model classification accuracy, providing a deeper understanding of how models comprehend and process visual prompts.

A different approach does not superimpose visual prompts directly onto images but treats visual prompts as an additional and independent learnable component \cite{wu2022unleashing}. It enhances model accuracy and adaptability in complex visual tasks by appropriately scaling images to coordinate their pixel ratio with visual prompts. The paper also indicates that proper normalization can improve the performance of visual prompting.

\subsection{Finetuneing Methods}

Efficiently finetuning pretrained models for specific tasks is an attractive area of research in CV. Currently, despite full finetune, mainstream efficient finetuning techniques can be categorized into two main types: methods that change the model structure and those that do not. Methods that alter the model structure include Adapter \cite{chen2022vision} and LoRA \cite{hu2021lora}. Full finetune is a traditional finetuning method that involves updating all parameters of the model. Adapter and LoRA are lightweight finetuning techniques that introduce a minimal number of additional parameters into the model. Without changing the original pretrained parameters of the model, only the newly introduced parameters are trained for finetuning. The distinction is that Adapter works in series with the backbone model, while LoRA operates in parallel.

As for techniques that do not change the model structure, the popular Linear Probe approach does not alter the model's architecture but adds a linear classification head after the pretrained model. It adapts to new tasks by swiftly learning the weights of the linear layer. Prompt engineering refers to changing the distribution of input data to enhance model performance in downstream tasks.

The choice of an appropriate finetuning strategy depends on specific task requirements, resource availability, and the similarity between the model and the task. From the comprehensive learning of full finetune to the parameter economy of LoRA and Adapter, and the structural preservation of LP and VPT \cite{jia2022visual}, each method offers different trade-offs in practical applications.

\section{Method}
Finding how to effectively couple the visual prompting and linear probe is the key idea of our method. We will elaborate our method and how to design the loss function to promote the collaboration. In section 3.1, we define the optimization problem and mathematical symbols. Then in section 3.2, we conclude the properties of VP and LP, and their limitations. For section 3.3, details of MoVL (Mixture of Visual Prompting and Linear Probe) will be presented.  

\begin{figure}[h]
    \centering
    \begin{subfigure}{0.23\textwidth}
        \centering
        \includegraphics[width=\linewidth]{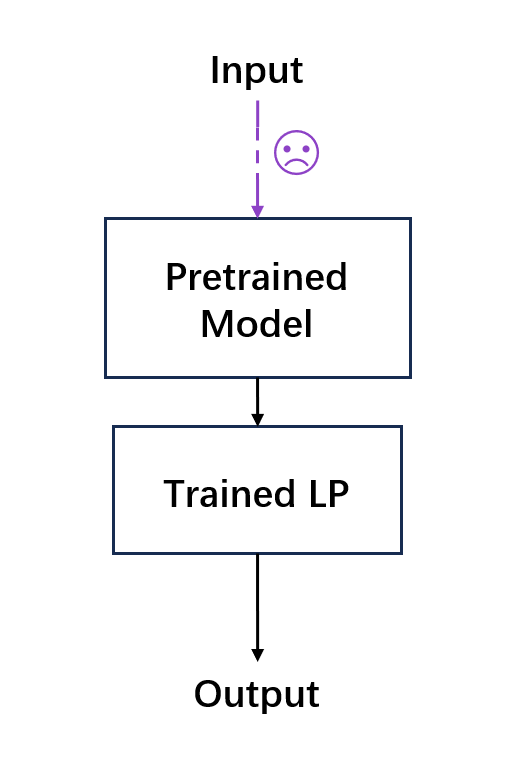}
        \caption{}
    \end{subfigure}
    \begin{subfigure}{0.23\textwidth}
        \centering
        \includegraphics[width=\linewidth]{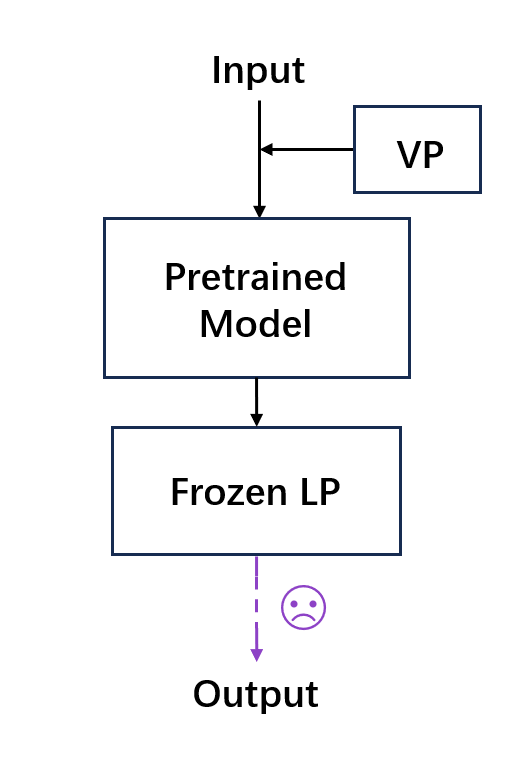}
        \caption{}
    \end{subfigure}
    \caption{Limitations of LP and VP. (a) shows the gap from input medical images to natural pretrained model; (b) shows the gap from output labels to specific ground truth labels;}
    \label{figure2}
\end{figure}

\begin{figure*}[t]
  \centering
  \includegraphics[width=\linewidth]{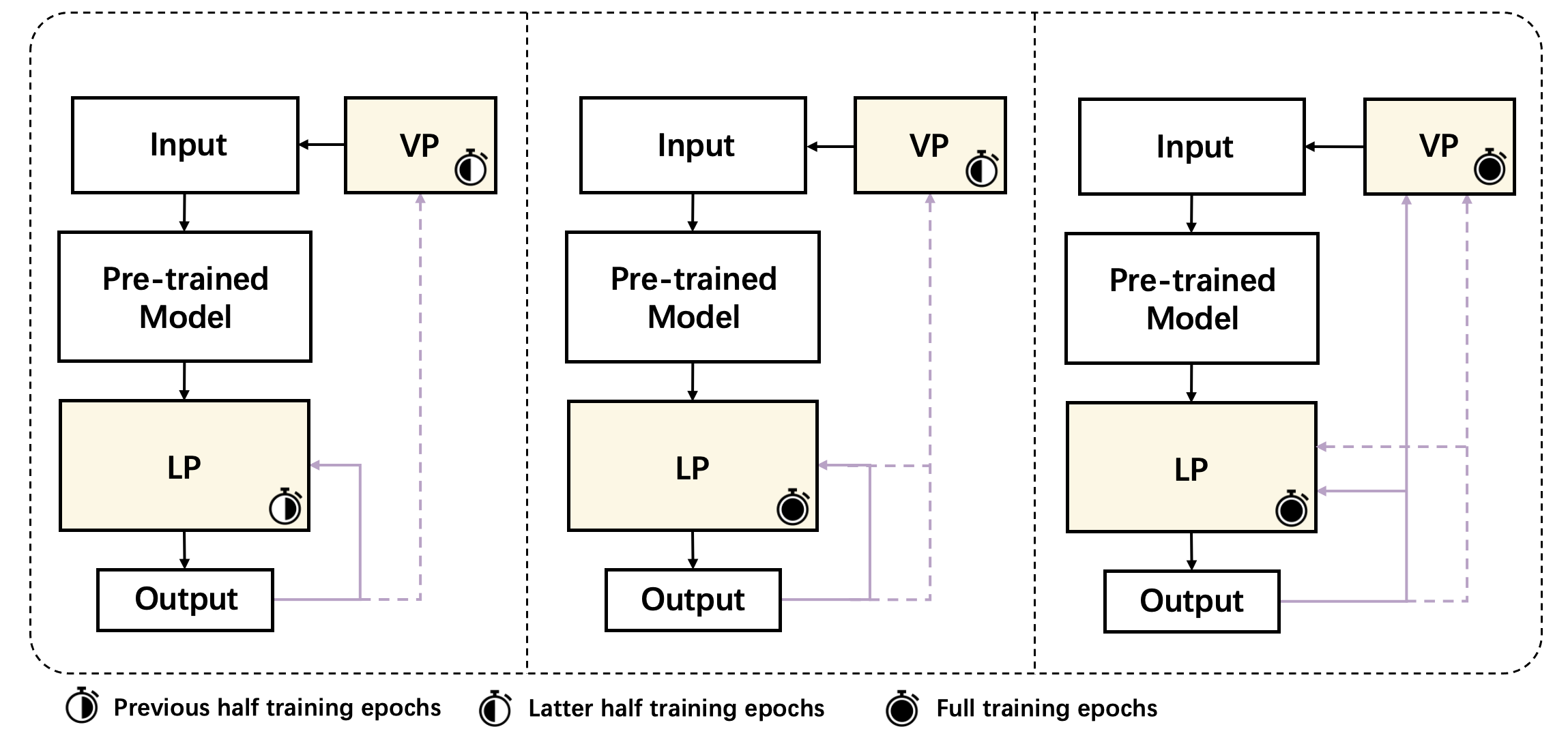}
  \caption{Three different training strategies. The left shows that first train LP and then training VP; The middle shows that first train LP and then training LP and VP together; The right show that train LP and VP together during the full training epochs.}
  \label{figure3}
\end{figure*}

\subsection{Problem Definition}
In this paper, the primary concern on determining the most effective approach for leveraging natural pretrained models to achieve accurate classification of medical images. Input images $x_T$ are randomly selected from the target medical domain dataset $D_T$. Subsequently, we add VP to the input image. This process can be denoted as:$$\hat{x_T}=x_T\oplus\delta$$
where $\hat{x_T}$ denotes the prompted images, $\delta$ denotes VP and $\oplus$ denotes different ways of adding VP, which can create a pixel patch at random or fix location, paddling around the images \cite{bahng2022exploring} or resize the input images first and then paddling around the images \cite{wu2022unleashing}.
Next, we input $\hat{x_T}$ into the pretrained backbone model $f$, and used the classifier $c$ to get final result. In this process, we formulate the optimization object:
$$Minimize \mathbf{E}_{(x_T,y_T)\in D_{T_{Tr}}}[l(c(f_{\theta_S}(\hat{x_T})),y_T)]$$
where $l$ denotes loss function, $D_{T_{Tr}}$ denotes training dataset of target medical domain, and $\theta_S$ denotes pretrained backbone model parameters.





\subsection{Comparison with LP and VP}

For transfer learning with only LP, the input can be rewrited as 

$$\hat{x_T}=x_T$$

There is no visual prompt added to the input image and vanilla medical images will be sent to natural image encoder. We notice from $x_T$ to $\theta_S$ existing a gap from target to source domain. For example, ResNet \cite{he2016deep} is pretrained on imagenet \cite{deng2009imagenet}, which is a dataset of natural images. $\theta_S$ of ResNet is a set of particular parameters to extract the natural features. Despite medical images may contains some of the common features, but still maintains unique medical features.

For transfer learning with only VP, the formula can be rewrited as:

$$Minimize \mathbf{E}_{(x_T,y_T)\in D_{T_{Tr}}}[l(c_{LM}(f_{\theta_S}(\hat{x_T})),y_T)]$$

$c_{LM}$ denotes the classifier for label matching (LM). Random label matching (RLM) and frequency label matching (FLM) are usually used in VP label matching process. RLM appoint the output label randomly, without considering the relationship between the labels. FLM select labels according to the probability of the labels, but without considering the change of VP. Although there is already work \cite{chen2023understanding} researching on LM, the accuracy of VP is still hard to reach LP's.

\subsection{MoVL}

%
MoVL (\textbf{M}ixture \textbf{o}f \textbf{V}isual Prompting and \textbf{L}inear Probe) is an designed algorithm containing training strategy and joint learning loss. The algorithm process is following:

\begin{algorithm}
    \caption{MoVL training process}
    \label{alg:algorithm}
    \textbf{Input}: Target domain images $x_T$\\
    \textbf{Parameter}: Frozen backbone image encoder parameters $\theta_S$, unfrozen VP parameters $\theta_{VP}$, unfrozen LP parameters $\theta_{LP}$, visual prompting $\delta$, linear probe $c()$, image encoder $f()$, $LS()$ denotes for correct label confidence selecting \\
    \begin{algorithmic}[1] 
        \FOR {Epochs n = 0,1,...}  
        \STATE \textbf{Forward process:}
            \STATE\hspace*{1em}$\hat{x_T}=x_T\oplus\delta$
            \STATE\hspace*{1em}$p_{-VP} = LS(Softmax(c(f_{\theta_S}(x_T))))$
            \STATE \hspace*{1em}$p_{+VP} = LS(Softmax(c(f_{\theta_S}(\hat{x_T}))))$
            \STATE\hspace*{1em}$l=(1-\alpha(p_{+VP}-p_{-VP}))CE$
        \STATE \textbf{Back propagation process:}
            \STATE\hspace*{1em}Update $\theta_{VP}$ and $\theta_{LP}$
        \ENDFOR
    \end{algorithmic}
    \textbf{return}
\end{algorithm}

\subsubsection{Training strategy}
Taking the properties of LP and VP into account, it is complementary and compatible for two parts, but how to couple the two parts still pends. We design three different training strategies, LP$\rightarrow$VP, LP$\rightarrow$mix, and  mix, as shown in Figure \ref{figure3}. The three training strategies are as follows:

$\bullet$ LP$\rightarrow$VP: First train LP and then VP.

$\bullet$ LP$\rightarrow$mix: First train LP and then together LP and VP.

$\bullet$ mix: Train LP and VP from the beginning.

Noting that LP is always trained first or together with VP, for the label matching deficiency of VP. If VP is trained first, the learned visual prompt corresponding to wrong label, VP may not contribute to the correctly matched label. So we trained LP for correct label matching first, and then VP to fill the gap between source natural images to target medical images. Additionally, mix strategy may help with each module, because VP changes the distribution of input space and LP classify the feature map to corresponding labels, which is the output space. When input space and output space are both changeable, it is an intuitive way to train two parts together with frozen parameters $\theta_S$. 

Further considering LP$\rightarrow$VP strategy, if LP is frozen when VP trained, although LP can match labels to some extent, VP is constrained to locked LP. Essentially, VP still confront LM problem in section 3.2. However, mix strategy can alleviate the problem, because both VP and LP are trained in a conjunct space and finally converge to global optimization together, which may avoid falling into a local optimum. So we select mix as a base strategy.
 
\subsubsection{Joint learning loss}

To enhance the training of LP and VP by utilizing the information from the original input images, we specially designed the loss function. Many previous study, such as \cite{wu2022unleashing}, \cite{chen2023understanding}, \cite{chen2022vision} may only use images with VP for training, ignoring the information containing in original input images. Thus we redesign the loss function to make use of neglected information from original input. Specifically, the loss function can be represented as: 
$$l=(1-\alpha(p_{+VP}-p_{-VP}))CE$$ 
$p_{+VP}$ denotes the correct classification probability of input image with VP and $p_{-VP}$ denotes the correct classification probability of input image without VP. $CE$ is the vanilla cross entropy loss, which is also called categorisation loss. The difference of $p_{+VP}$ and $p_{-VP}$ depict how far prompted image exceed reference, the original image without prompt, which is named as discrepancy loss. Joint learning loss function combines categorisation loss and discrepancy loss and $\alpha$ denotes a superparameter to balance two parts weights. 

From intuitive understanding, we encourage the model improve the confidence of correct classification with VP contrast to without VP, we name this 'confidence improvement'. We take the results of images without VP as a baseline, to instruct how further the results of images with VP can be better. The more $p_{+VP}$ beyonds $p_{-VP}$, the less of the joint learning loss value. 

There need to be highlighted that with only such a loss can not achieve the goal of improving confidence of VP and get a better accuracy, for $p_{+VP}-p_{-VP}$ can be widen by suppressing $p_{-VP}$, which may just decrease the confidence of original results. Hence, we detach the original output results to avoid such suppression, with only update the parameters of with VP process.

\section{Experiments}

In this section, we will demonstrate the experiment details and the results among different datasets and baselines, which followed by analysis and explanation.

\subsection{Experiment setups}

We conducted the experiments with four datasets, inclulding HAM10000\cite{tschandl2018ham10000}, PathMNIST, BloodMNIST\cite{medmnistv2,medmnistv1}, and Camelyon17 \cite{bandi2018detection}.
HAM10000 \cite{tschandl2018ham10000} is a dataset of skin lesion images, which contains a total of 10,015 images of seven types of skin lesions from clinical settings: actinic keratoses, basal cell carcinoma, bening keratosis-like lesions, dermatofibroma, melanocytic nevi, vascular lesions, and melanoma. 

PathMNIST and BloodMNIST are two tasks in MedMNIST for the classification of biomedical images.\cite{medmnistv2,medmnistv1}. PathMNIST is a dataset collected from colorectal cancer histology slides, which contains 107,180. There are nine types of tissues in PathMNIST: adipose, background, debris, lymphocytes, mucus, smooth muscle, normal colon mucosa, cancer-associated stroma, colorectal adenocarscinoma epithelium. BloodMNIST contains images of normal cells without infection, hematologic or oncologic disease, and free of any pharmacologic treatment at the time of blood collection. It consists of a total of 17,092 images categorized into 8 classes, which are basophil, eosinophil, erythroblast, immature granulocytes, lymphocyte, monocyte, neutrophil, and platelet.

Camelyon17 \cite{bandi2018detection} is a dataset designed to probe model's capability on the condition of out-of-distribution (OOD). We use the version from WILDS benchmark \cite{wilds2021,sagawa2022extending}. The tissue patches are collected from different hospitals, then was divided into training, validation and test set, which is a classic circumstance of distribution shift. It contains 45K images of data with two categories, benign or tumor.

The official split is adopted for all datasets. 

\begin{table*}[t]
  \centering
  \begin{subtable}{.47\textwidth}
    \centering
    \begin{tabular}{|c|c|c|c|c|}
      \hline
      Baseline & ResNet18 & ResNet50 & ViT-B/32 &  ViT-L/14\\
      \hline
      LP & 82.028 & 92.362 & 83.712 & 88.303 \\
      VP & 73.921 & 77.224 & 84.351 & 95.100 \\
      MoVL & \textbf{91.648} & \textbf{94.460} & \textbf{94.066} & 91.339\\
      \rowcolor[gray]{0.9}FF & 99.084 & 98.913 & 99.265 & 99.425\\
      \hline
    \end{tabular}
    \caption{HAM10000}
  \end{subtable}
  \hfill
  \begin{subtable}{.47\textwidth}
    \centering
    \begin{tabular}{|c|c|c|c|c|}
      \hline
      Baseline & ResNet18 & ResNet50 & ViT-B/32 &  ViT-L/14\\
      \hline
      LP & 86.532 & 88.872 & 89.123 & 89.457 \\
      VP & 74.624 & 77.145 & 81.518 & 87.173 \\
      MoVL & \textbf{87.130} & \textbf{89.540} & \textbf{92.033} & \textbf{90.585}\\
      \rowcolor[gray]{0.9}FF & 91.365 & 86.797 & 93.370 & 94.875\\
      \hline
    \end{tabular}
    \caption{PathMNIST}
  \end{subtable}
  
  \medskip

  \begin{subtable}{.47\textwidth}
    \centering
    \begin{tabular}{|c|c|c|c|c|}
      \hline
      Baseline & ResNet18 & ResNet50 & ViT-B/32 &  ViT-L/14\\
      \hline
      LP & 87.284 & 89.652 & 88.395 & 90.119 \\
      VP & 73.020 & 73.517 & 77.580 & 85.443 \\
      MoVL & \textbf{90.909} & \textbf{91.611} & \textbf{90.792} & \textbf{92.692} \\
      \rowcolor[gray]{0.9}FF & 96.200 & 96.083 & 96.638 & 98.100\\
      \hline
    \end{tabular}
    \caption{BloodMNIST}
  \end{subtable}
  \hfill
  \begin{subtable}{.47\textwidth}
    \centering
    \begin{tabular}{|c|c|c|c|c|}
      \hline
      Baseline & ResNet18 & ResNet50 & ViT-B/32 &  ViT-L/14\\
      \hline
      LP & 87.657 & 89.373 & 90.331 & 89.666 \\
      VP & 86.271 & 85.411 & 86.059 & 89.772\\
      MoVL & \textbf{89.091} & \textbf{89.675} & \textbf{92.075} & \textbf{90.462} \\
      \rowcolor[gray]{0.9}FF & 80.985 & 79.446 & 86.861 & 93.299\\
      \hline
    \end{tabular}
    \caption{Camelyon17*}
  \end{subtable}
  \caption{Comparison with other baselines on four different datasets. HAM10000, PathMNIST and BloodMNIST are ID datasets and Camelyon17 is OOD dataset, which is marked by *. We use ResNet18, ResNet50, ViT-B/32 and ViT-L/14 to experiment. We use full finetune result as an upper bound on ID datasets. But for OOD dataset, MoVL can outperform full finetune (FF).}
  \label{table1}
\end{table*}

We conducted the experiments with ResNet \cite{he2016deep} and CLIP \cite{radford2021learning} as backbone models. For each of the architecture, we select two different sizes to show the robustness of our method. For ResNet, we choose ResNet18 and ResNet50, which are pretrained on Imagenet-1K \cite{deng2009imagenet}. For CLIP, we choose ViT-B/32 and ViT-L/14 visual encoder.

We train the model with AdamW optimizer \cite{loshchilov2017decoupled}. Except for mix strategy of ViT-L/14, we use the learning rate of 0.01. Because we find that it is hard for mix strategy convergence with such big backbone model as the same epochs with other model settings, we modify the learning rate to 0.1 for all mix strategy experiments with ViT-L/14. We use cosine scheduler \cite{loshchilov2016sgdr} to decrease the learning rate as epochs increasing and weight decay is set to 0. Other parameters are default for AdamW optimizer. Warmup steps is set to 10. For HAM10000, PathMNIST and BloodMNIST datasets, we use the batch size of 128. For Camelyon17 dataset, we use the batchsize of 256. For all experiments, we use the prompt method of paddle and the prompt size is set to 30 pixels. The input images are first resize to (224-2*30)*(224-2*30) and then adding VP, so the prompted input images are 224*224 size \cite{wu2022unleashing}. VP is initialized as all zeroes pixels with minimum $\epsilon$ = 0.001.


For only LP or mix training period, namely one stage training, we use 20 epochs totally. For two stage training, we train 10 epochs for each stage.

\subsection{Experiment Results}

For model validation, we first choose full finetune as the upper benchmark on ID datasets, given its capability to update a larger number of parameters. We also choose LP and VP as two baseline finetuning methods. These methods do not change the original parameters or architecture of the backbone image encoder. For LP and VP, we adhered to the same settings mentioned in section 4.1. For full finetune, we still use learning rate 0.01 for ResNet but 0.0001 for CLIP. The text prompt for CLIP follows the work by\cite{radford2021learning},specifically 'This is a photo of a \{ \}'. For dataset Camelyon17, we use text labels 'benign tissue region' and 'tissue tumor region'. For other datasets, we use the labels as described in section 4.1.

From Table \ref{table1}, VP performs better using CLIP architecture. This may be caused by output label matching gap of frozen linear probe in ResNet. For text prompt of CLIP, calculating the similarity between text and image can remit the gap to some extent, but professional medical terms can still be awkward. LP performs relatively stable than VP. However, there is a great distance from full finetune, which may caused by medical input and natural image encoder gap.

Our proposed MoVL (average of \textbf{90.917}\%) can outperform than single LP (average of \textbf{88.304}\%) or VP (average of \textbf{81.758}\%) for most experiments, indicating that only one module involved can be restricted by its proverties. However, a mixture strategy can be complementary and improve the accuracy. When contrast with full finetune (average of \textbf{91.132}\%), there is potential for MoVL to reach accuracy of full finetune.

We notice that on OOD dataset Camelyon17, full finetune can get worse performance than other methods. This is consistent with prior work \cite{kumar2022fine}. There is a gap between ID and OOD data, and full finetune only fit with ID, but distort OOD features. However, MoVL can be less influenced by this distribution shift and take \textbf{5.178}\% lead.

We observe that for ViT-L/14, some resluts of MoVL are lower than ViT-B/32, which may be caused by more training epochs to get convergence for larger backbone model, and we use the same consistent epochs for different models. 

We observe MoVL (average of \textbf{0.029M}) use far less parameters compared with full finetune (average of \textbf{107.3M}), with the sum parameters of LP and VP. Parameters of VP is the same because of the same prompt size and input image size, but LP will change with feature map channel and number of categories.

\subsection{Strategy Validation Experiments}

In this section, we do experiments with different training strategies: LP only, LP$\rightarrow$VP, LP$\rightarrow$mix and mix all time. We choose LP only as a baseline to analyze, for LP is the base of other strategies. We also use various backbone models to show the efficiency of our proposed method. Both ID and OOD datasets are used to show the validation experiment results. 

\begin{table}[h!]
\centering
\begin{subtable}[h]{0.49\textwidth}
\centering
\begin{tabular}{lllll}
\toprule
Strategy &   ResNet18 &   ResNet50 &    ViT-B/32 &    ViT-L/14\\
\midrule
        LP &     82.028 &     92.362 &     83.712 &     88.303 \\
     LP$\rightarrow$VP &     80.121 &      84.510 &     85.821 &     74.241 \\
    LP$\rightarrow$mix &     87.834 &     92.298 &     90.796 &     90.210 \\
       mix &     \textbf{91.467} &     \textbf{94.439} &     \textbf{93.704} &     \textbf{91.787} \\
\bottomrule
\end{tabular}
\caption{HAM10000}
\end{subtable}
\begin{subtable}[h]{0.49\textwidth}
\centering
\begin{tabular}{lllll}
\toprule
Strategy &   ResNet18 &   ResNet50 &    ViT-B/32 &    ViT-L/14\\
\midrule
        LP &     86.532 &     88.872 &     89.123 &     89.457 \\
     LP$\rightarrow$VP &     85.404 &     87.535 &     90.348 &     87.103 \\
    LP$\rightarrow$mix &     86.769 &     89.095 &     90.738 &     90.139 \\
       mix &     \textbf{86.963} &     \textbf{89.428} &     \textbf{91.629} &     \textbf{90.446} \\
\bottomrule
\end{tabular}
\caption{PathMNIST}
\end{subtable}
\begin{subtable}[h]{0.49\textwidth}
\centering
\begin{tabular}{lllll}
\toprule
Strategy &   ResNet18 &   ResNet50 &    ViT-B/32 &    ViT-L/14\\
\midrule
        LP &     87.284 &     89.652 &     88.395 &     90.119 \\
     LP$\rightarrow$VP &     86.524 &     87.518 &     87.138 &     86.027 \\
    LP$\rightarrow$mix &     89.594 &     91.055 &     89.068 &     90.880 \\
       mix &     \textbf{90.558} &     \textbf{91.464} &       \textbf{90.500} &     \textbf{91.815} \\
\bottomrule
\end{tabular}
\caption{BloodMNIST}
\end{subtable}
\begin{subtable}[h]{0.49\textwidth}
\centering
\begin{tabular}{lllll}
\toprule
Strategy &   ResNet18 &   ResNet50 &    ViT-B/32 &    ViT-L/14\\
\midrule
        LP &     87.657 &     89.373 &     90.331 &     89.666 \\
     LP$\rightarrow$VP &     87.844 &     88.671 &     91.133 &     88.133 \\
    LP$\rightarrow$mix &     88.583 &     \textbf{89.459} &     91.938 &     \textbf{91.849} \\
       mix &     \textbf{89.044} &     89.445 &      \textbf{92.050} &     \textbf{91.849} \\

\bottomrule
\end{tabular}
\caption{Camelyon17*}
\end{subtable}

\caption{Different training strategy experiments on three ID datasets and one OOD datasets. * denotes OOD dataset. LP$\rightarrow$VP denotes for first LP training and then VP training; LP$\rightarrow$mix denotes for first LP training and then LP and VP training together.}
\label{table2}
\end{table}

From Table \ref{table2}, we observe adding VP may not be outperform than train LP for the full time, which indicates that adding modules in arbitrary combination  doesn't guarantee enhanced accuracy.

Under most circumstances, use mix strategy can improve the accuracy of categorizing tasks, no matter after LP training or mix training from the beginning, which indicates validity of mix strategy. There is also a trend that with more epochs of LP and VP training together, the better performance of the result, which indicates that both input and output space trainable is helpful with the improvement of the accuracy.

We also do the strategy analysis on OOD dataset, Camelyon17. The conclusion of ID dataset can still work on OOD dataset, but with a discount on improvement. This can be explained by the distribution shift from training and testing dataset. Nonetheless, mix strategy can help with combining the benefits of LP and VP.

\begin{table}[h!]
\centering
\begin{tabular}{lllll}
\toprule
Strategy &   ResNet18 &   ResNet50 &    ViT-B/32 &    ViT-L/14\\
\midrule
        LP &     85.281 &     90.295 &     87.077 &     89.293 \\
     LP$\rightarrow$VP &     86.166 &     89.107 &     88.797 &     81.514 \\
    LP$\rightarrow$mix &     89.072 &     91.918 &     90.535 &     89.837 \\
       mix &     \textbf{90.063} &     \textbf{92.188} &     \textbf{91.783} &     \textbf{91.050} \\

\bottomrule
\end{tabular}
\caption{ The average accuracy of another visual prompt method on three ID datasets.}
\label{table3}
\end{table}

To prove the training strategy is efficient for different situations, we change the visual prompt method. Following the workflow of \cite{bahng2022exploring}, we directly add VP onto the original input images, which may cover the border of the images. The results still shows that mix strategy can be more competitive than other training strategies.

\subsection{Loss Ablation Experiments}

From the experiments above, we have shown that mix strategy can be quite a competitive strategy combining LP and VP. In this section, we will interpret our designed joint learning loss function can make use of information from original output without prompt to improve the accuracy of the model.

\begin{table}[h!]
\centering
\begin{tabular}{lllll}
\toprule
&   ResNet18 &   ResNet50 &    ViT-B/32 &    ViT-L/14\\
\midrule
  HAM10000 &    +0.181 &    +0.021 &    +0.362 &     -0.448 \\
 PathMNIST &    +0.167 &    +0.112 &    +0.404 &    +0.139 \\
BloodMNIST &    +0.351 &    +0.147 &    +0.292 &    +0.887 \\
Camelyon17*&    +0.047 &    +0.230 &     +0.025 &     -1.387 \\
\bottomrule
\end{tabular}
\caption{With designed joint learning loss, mix strategy can perform better than just cross entropy loss in most cases}
\label{table5}
\end{table}

We counted data from different models on different datasets and found that using the joint learning loss function can improve the accuracy of the models. This may because cross entropy loss can only get the information after adding VP, with only one forward propagation. However, as original images getting through image encoder and LP, then gaining the original output, which can still contains information. For example, an original dog image can be wrongly categorized as cat, but this information can be referred by prompted one, which is not cat. As the model classify the image correctly, it can still be a reference for prompted one to exceed. On considering such information, VP can reach and beyond reference output.

\subsection{Exploration of VP Initialization}

\begin{table}[h!]
\centering
\begin{tabular}{lllll}
\toprule
&   ResNet18 &   ResNet50 &   \\
\midrule
  HAM10000 &     89.634 (-2.014) &     92.543 (-1.917) \\
 PathMNIST &     85.598 (-1.532) &     88.510 (-1.030) \\
BloodMNIST &     90.558 (-0.351) &   91.318 (-0.293)  \\
\bottomrule
\end{tabular}
\caption{Random initialization contrast with all zeroes plus a minimum $\epsilon$ = 0.001 initialization}
\label{table7}
\end{table}

We found initialization of VP can greatly influence the accuracy. With a simple 'blank' VP, which is all zeroes plus a minimum $\epsilon$ = 0.001 initialization, can be better than random initialization.

\section{Conclusion and Future Work}

This paper explore the proper strategy for LP and VP combining, which is inspired by the properties of LP and VP. Training LP and VP together from the beginning outperforms than other strategies. Besides, we propose a joint learning loss function, which contains categorisation loss and discrepancy loss. Discrepancy loss takes information of the original imageinto consideration. We do experiments on different backbone models, both ID and OOD distribution datasets. The results shows the efficiency of our method.

We regard the MoVL as a promising strategy, which can adapt to different downstream tasks with a huge gap from natural. The idea of referring to original image may guide more delicate loss design about VP training. Initialization of VP is also very important for training process, and we will further explore it in the future work.

\bibliographystyle{named}
\bibliography{ijcai24}

\begin{thebibliography}{}

\bibitem[\protect\citeauthoryear{Anonymous}{2023}]{anonymous2023visual}
Anonymous.
\newblock Visual prompting reimagined: The power of activation prompts.
\newblock In {\em Submitted to The Twelfth International Conference on Learning Representations}, 2023.
\newblock under review.

\bibitem[\protect\citeauthoryear{Bahng \bgroup \em et al.\egroup }{2022}]{bahng2022exploring}
Hyojin Bahng, Ali Jahanian, Swami Sankaranarayanan, and Phillip Isola.
\newblock Exploring visual prompts for adapting large-scale models.
\newblock {\em arXiv preprint arXiv:2203.17274}, 2022.

\bibitem[\protect\citeauthoryear{Bandi \bgroup \em et al.\egroup }{2018}]{bandi2018detection}
Peter Bandi, Oscar Geessink, Quirine Manson, Marcory Van~Dijk, Maschenka Balkenhol, Meyke Hermsen, Babak~Ehteshami Bejnordi, Byungjae Lee, Kyunghyun Paeng, Aoxiao Zhong, et~al.
\newblock From detection of individual metastases to classification of lymph node status at the patient level: the camelyon17 challenge.
\newblock {\em IEEE transactions on medical imaging}, 38(2):550--560, 2018.

\bibitem[\protect\citeauthoryear{Brown \bgroup \em et al.\egroup }{2020}]{brown2020language}
Tom Brown, Benjamin Mann, Nick Ryder, Melanie Subbiah, Jared~D Kaplan, Prafulla Dhariwal, Arvind Neelakantan, Pranav Shyam, Girish Sastry, Amanda Askell, et~al.
\newblock Language models are few-shot learners.
\newblock {\em Advances in neural information processing systems}, 33:1877--1901, 2020.

\bibitem[\protect\citeauthoryear{Chen \bgroup \em et al.\egroup }{2022}]{chen2022vision}
Zhe Chen, Yuchen Duan, Wenhai Wang, Junjun He, Tong Lu, Jifeng Dai, and Yu~Qiao.
\newblock Vision transformer adapter for dense predictions.
\newblock {\em arXiv preprint arXiv:2205.08534}, 2022.

\bibitem[\protect\citeauthoryear{Chen \bgroup \em et al.\egroup }{2023}]{chen2023understanding}
Aochuan Chen, Yuguang Yao, Pin-Yu Chen, Yihua Zhang, and Sijia Liu.
\newblock Understanding and improving visual prompting: A label-mapping perspective.
\newblock In {\em Proceedings of the IEEE/CVF Conference on Computer Vision and Pattern Recognition}, pages 19133--19143, 2023.

\bibitem[\protect\citeauthoryear{Chen \bgroup \em et al.\egroup }{2024}]{chen2024eliciting}
Yadang Chen, Gang Yang, Duolin Wang, and Dichao Li.
\newblock Eliciting knowledge from language models with automatically generated continuous prompts.
\newblock {\em Expert Systems with Applications}, 239:122327, 2024.

\bibitem[\protect\citeauthoryear{Deng \bgroup \em et al.\egroup }{2009}]{deng2009imagenet}
Jia Deng, Wei Dong, Richard Socher, Li-Jia Li, Kai Li, and Li~Fei-Fei.
\newblock Imagenet: A large-scale hierarchical image database.
\newblock In {\em 2009 IEEE conference on computer vision and pattern recognition}, pages 248--255. Ieee, 2009.

\bibitem[\protect\citeauthoryear{Gu \bgroup \em et al.\egroup }{2023}]{gu2023systematic}
Jindong Gu, Zhen Han, Shuo Chen, Ahmad Beirami, Bailan He, Gengyuan Zhang, Ruotong Liao, Yao Qin, Volker Tresp, and Philip Torr.
\newblock A systematic survey of prompt engineering on vision-language foundation models.
\newblock {\em arXiv preprint arXiv:2307.12980}, 2023.

\bibitem[\protect\citeauthoryear{Hambardzumyan \bgroup \em et al.\egroup }{2021}]{hambardzumyan2021warp}
Karen Hambardzumyan, Hrant Khachatrian, and Jonathan May.
\newblock Warp: Word-level adversarial reprogramming.
\newblock {\em arXiv preprint arXiv:2101.00121}, 2021.

\bibitem[\protect\citeauthoryear{He \bgroup \em et al.\egroup }{2016}]{he2016deep}
Kaiming He, Xiangyu Zhang, Shaoqing Ren, and Jian Sun.
\newblock Deep residual learning for image recognition.
\newblock In {\em Proceedings of the IEEE conference on computer vision and pattern recognition}, pages 770--778, 2016.

\bibitem[\protect\citeauthoryear{Hu \bgroup \em et al.\egroup }{2021}]{hu2021lora}
Edward~J Hu, Yelong Shen, Phillip Wallis, Zeyuan Allen-Zhu, Yuanzhi Li, Shean Wang, Lu~Wang, and Weizhu Chen.
\newblock Lora: Low-rank adaptation of large language models.
\newblock {\em arXiv preprint arXiv:2106.09685}, 2021.

\bibitem[\protect\citeauthoryear{Jia \bgroup \em et al.\egroup }{2022}]{jia2022visual}
Menglin Jia, Luming Tang, Bor-Chun Chen, Claire Cardie, Serge Belongie, Bharath Hariharan, and Ser-Nam Lim.
\newblock Visual prompt tuning.
\newblock In {\em European Conference on Computer Vision}, pages 709--727. Springer, 2022.

\bibitem[\protect\citeauthoryear{Kirillov \bgroup \em et al.\egroup }{2023}]{kirillov2023segment}
Alexander Kirillov, Eric Mintun, Nikhila Ravi, Hanzi Mao, Chloe Rolland, Laura Gustafson, Tete Xiao, Spencer Whitehead, Alexander~C Berg, Wan-Yen Lo, et~al.
\newblock Segment anything.
\newblock {\em arXiv preprint arXiv:2304.02643}, 2023.

\bibitem[\protect\citeauthoryear{Koh \bgroup \em et al.\egroup }{2021}]{wilds2021}
Pang~Wei Koh, Shiori Sagawa, Henrik Marklund, Sang~Michael Xie, Marvin Zhang, Akshay Balsubramani, Weihua Hu, Michihiro Yasunaga, Richard~Lanas Phillips, Irena Gao, Tony Lee, Etienne David, Ian Stavness, Wei Guo, Berton~A. Earnshaw, Imran~S. Haque, Sara Beery, Jure Leskovec, Anshul Kundaje, Emma Pierson, Sergey Levine, Chelsea Finn, and Percy Liang.
\newblock {WILDS}: A benchmark of in-the-wild distribution shifts.
\newblock In {\em International Conference on Machine Learning (ICML)}, 2021.

\bibitem[\protect\citeauthoryear{Krizhevsky \bgroup \em et al.\egroup }{2009}]{krizhevsky2009learning}
Alex Krizhevsky, Geoffrey Hinton, et~al.
\newblock Learning multiple layers of features from tiny images.
\newblock 2009.

\bibitem[\protect\citeauthoryear{Kumar \bgroup \em et al.\egroup }{2022}]{kumar2022fine}
Ananya Kumar, Aditi Raghunathan, Robbie Jones, Tengyu Ma, and Percy Liang.
\newblock Fine-tuning can distort pretrained features and underperform out-of-distribution.
\newblock {\em arXiv preprint arXiv:2202.10054}, 2022.

\bibitem[\protect\citeauthoryear{Liu \bgroup \em et al.\egroup }{2023}]{liu2023pre}
Pengfei Liu, Weizhe Yuan, Jinlan Fu, Zhengbao Jiang, Hiroaki Hayashi, and Graham Neubig.
\newblock Pre-train, prompt, and predict: A systematic survey of prompting methods in natural language processing.
\newblock {\em ACM Computing Surveys}, 55(9):1--35, 2023.

\bibitem[\protect\citeauthoryear{Loshchilov and Hutter}{2016}]{loshchilov2016sgdr}
Ilya Loshchilov and Frank Hutter.
\newblock Sgdr: Stochastic gradient descent with warm restarts.
\newblock {\em arXiv preprint arXiv:1608.03983}, 2016.

\bibitem[\protect\citeauthoryear{Loshchilov and Hutter}{2017}]{loshchilov2017decoupled}
Ilya Loshchilov and Frank Hutter.
\newblock Decoupled weight decay regularization.
\newblock {\em arXiv preprint arXiv:1711.05101}, 2017.

\bibitem[\protect\citeauthoryear{Petroni \bgroup \em et al.\egroup }{2019}]{petroni2019language}
Fabio Petroni, Tim Rockt{\"a}schel, Patrick Lewis, Anton Bakhtin, Yuxiang Wu, Alexander~H Miller, and Sebastian Riedel.
\newblock Language models as knowledge bases?
\newblock {\em arXiv preprint arXiv:1909.01066}, 2019.

\bibitem[\protect\citeauthoryear{Radford \bgroup \em et al.\egroup }{2021}]{radford2021learning}
Alec Radford, Jong~Wook Kim, Chris Hallacy, Aditya Ramesh, Gabriel Goh, Sandhini Agarwal, Girish Sastry, Amanda Askell, Pamela Mishkin, Jack Clark, et~al.
\newblock Learning transferable visual models from natural language supervision.
\newblock In {\em International conference on machine learning}, pages 8748--8763. PMLR, 2021.

\bibitem[\protect\citeauthoryear{Sagawa \bgroup \em et al.\egroup }{2022}]{sagawa2022extending}
Shiori Sagawa, Pang~Wei Koh, Tony Lee, Irena Gao, Sang~Michael Xie, Kendrick Shen, Ananya Kumar, Weihua Hu, Michihiro Yasunaga, Henrik Marklund, Sara Beery, Etienne David, Ian Stavness, Wei Guo, Jure Leskovec, Kate Saenko, Tatsunori Hashimoto, Sergey Levine, Chelsea Finn, and Percy Liang.
\newblock Extending the wilds benchmark for unsupervised adaptation.
\newblock In {\em International Conference on Learning Representations (ICLR)}, 2022.

\bibitem[\protect\citeauthoryear{Tschandl \bgroup \em et al.\egroup }{2018}]{tschandl2018ham10000}
Philipp Tschandl, Cliff Rosendahl, and Harald Kittler.
\newblock The ham10000 dataset, a large collection of multi-source dermatoscopic images of common pigmented skin lesions.
\newblock {\em Scientific data}, 5(1):1--9, 2018.

\bibitem[\protect\citeauthoryear{Wang \bgroup \em et al.\egroup }{2022}]{wang2022medclip}
Zifeng Wang, Zhenbang Wu, Dinesh Agarwal, and Jimeng Sun.
\newblock Medclip: Contrastive learning from unpaired medical images and text.
\newblock {\em arXiv preprint arXiv:2210.10163}, 2022.

\bibitem[\protect\citeauthoryear{Wei \bgroup \em et al.\egroup }{2022}]{wei2022chain}
Jason Wei, Xuezhi Wang, Dale Schuurmans, Maarten Bosma, Fei Xia, Ed~Chi, Quoc~V Le, Denny Zhou, et~al.
\newblock Chain-of-thought prompting elicits reasoning in large language models.
\newblock {\em Advances in Neural Information Processing Systems}, 35:24824--24837, 2022.

\bibitem[\protect\citeauthoryear{Wu \bgroup \em et al.\egroup }{2022}]{wu2022unleashing}
Junyang Wu, Xianhang Li, Chen Wei, Huiyu Wang, Alan Yuille, Yuyin Zhou, and Cihang Xie.
\newblock Unleashing the power of visual prompting at the pixel level.
\newblock {\em arXiv preprint arXiv:2212.10556}, 2022.

\bibitem[\protect\citeauthoryear{Yang \bgroup \em et al.\egroup }{2021}]{medmnistv1}
Jiancheng Yang, Rui Shi, and Bingbing Ni.
\newblock Medmnist classification decathlon: A lightweight automl benchmark for medical image analysis.
\newblock In {\em IEEE 18th International Symposium on Biomedical Imaging (ISBI)}, pages 191--195, 2021.

\bibitem[\protect\citeauthoryear{Yang \bgroup \em et al.\egroup }{2023}]{medmnistv2}
Jiancheng Yang, Rui Shi, Donglai Wei, Zequan Liu, Lin Zhao, Bilian Ke, Hanspeter Pfister, and Bingbing Ni.
\newblock Medmnist v2-a large-scale lightweight benchmark for 2d and 3d biomedical image classification.
\newblock {\em Scientific Data}, 10(1):41, 2023.

\bibitem[\protect\citeauthoryear{Zhou \bgroup \em et al.\egroup }{2022a}]{zhou2022learning}
Kaiyang Zhou, Jingkang Yang, Chen~Change Loy, and Ziwei Liu.
\newblock Learning to prompt for vision-language models.
\newblock {\em International Journal of Computer Vision}, 130(9):2337--2348, 2022.

\bibitem[\protect\citeauthoryear{Zhou \bgroup \em et al.\egroup }{2022b}]{zhou2022self}
Lei Zhou, Huidong Liu, Joseph Bae, Junjun He, Dimitris Samaras, and Prateek Prasanna.
\newblock Self pre-training with masked autoencoders for medical image classification and segmentation.
\newblock {\em arXiv preprint arXiv:2203.05573}, 2022.

\end{thebibliography}

\end{document}